\documentclass[lettersize,journal]{IEEEtran}
\usepackage{amsmath,amsfonts}
\usepackage{algpseudocode}
\usepackage{algorithm}
\usepackage{array}
\usepackage[caption=false,font=normalsize,labelfont=sf,textfont=sf]{subfig}
\usepackage{textcomp}
\usepackage{stfloats}
\usepackage{url}
\usepackage{verbatim}
\usepackage{graphicx}
\usepackage{booktabs}
\usepackage{cite}
\usepackage{tabularx}
\usepackage{scalerel}
\usepackage{tikz}
\usepackage{flushend}
\usetikzlibrary{svg.path}
\definecolor{orcidlogocol}{HTML}{A6CE39}
\tikzset{
  orcidlogo/.pic={
    \fill[orcidlogocol] svg{M256,128c0,70.7-57.3,128-128,128C57.3,256,0,198.7,0,128C0,57.3,57.3,0,128,0C198.7,0,256,57.3,256,128z};
    \fill[white] svg{M86.3,186.2H70.9V79.1h15.4v48.4V186.2z}
                 svg{M108.9,79.1h41.6c39.6,0,57,28.3,57,53.6c0,27.5-21.5,53.6-56.8,53.6h-41.8V79.1z M124.3,172.4h24.5c34.9,0,42.9-26.5,42.9-39.7c0-21.5-13.7-39.7-43.7-39.7h-23.7V172.4z}
                 svg{M88.7,56.8c0,5.5-4.5,10.1-10.1,10.1c-5.6,0-10.1-4.6-10.1-10.1c0-5.6,4.5-10.1,10.1-10.1C84.2,46.7,88.7,51.3,88.7,56.8z};
  }
}

\newcommand\orcidicon[1]{\href{https://orcid.org/#1}{\mbox{\scalerel*{
\begin{tikzpicture}[yscale=-1,transform shape]
\pic{orcidlogo};
\end{tikzpicture}
}{|}}}}

\usepackage{hyperref} 

\begin{document}

\title{Neural Improvement Heuristics for \\ Graph Combinatorial Optimization Problems}

\author{Andoni I. Garmendia\orcidicon{0000-0002-7243-6116}, Josu Ceberio\orcidicon{0000-0001-7120-6338},~\IEEEmembership{Member,~IEEE,} Alexander Mendiburu\orcidicon{0000-0002-7271-1931},~\IEEEmembership{Member,~IEEE,}
\thanks{© 2023 IEEE. Personal use of this material is permitted. Permission
from IEEE must be obtained for all other uses, in any current or future
media, including reprinting/republishing this material for advertising or
promotional purposes, creating new collective works, for resale or
redistribution to servers or lists, or reuse of any copyrighted
component of this work in other works.}
}


\maketitle

\begin{abstract}

Recent advances in graph neural network architectures and increased computation power have revolutionized the field of combinatorial optimization (CO). Among the proposed models for CO problems, Neural Improvement (NI) models have been particularly successful. However, existing NI approaches are limited in their applicability to problems where crucial information is encoded in the edges, as they only consider node features and node-wise positional encodings. To overcome this limitation, we introduce a novel NI model capable of handling graph-based problems where information is encoded in the nodes, edges, or both. The presented model serves as a fundamental component for hill-climbing-based algorithms that guide the selection of neighborhood operations for each iteration. Conducted experiments demonstrate that the proposed model can recommend neighborhood operations that outperform conventional versions for the Preference Ranking Problem with a performance in the 99th percentile. We also extend the proposal to two well-known problems: the Traveling Salesman Problem and the Graph Partitioning Problem, recommending operations in the 98th and 97th percentile, respectively.

\end{abstract}


\section{Introduction}

Combinatorial Optimization Problems (COPs) are present in a broad range of real-world applications, such as logistics, manufacturing or biology \cite{paschos2014applications, naseri2020application}. Due to the \textit{NP-hard} nature of most COPs, finding the optimal solution applying exact methods becomes intractable as the size of the problem grows \cite{garey1979computers}. As a result, in the last few decades, heuristic and meta-heuristic methods have arisen as an alternative to approximate \textit{hard} COPs in a reasonable amount of time. Initial works in the field proposed constructive heuristic methods that iteratively build a candidate solution. In general, constructive methods are developed ad-hoc for the problem at hand, based on criteria and rules provided by expert knowledge. Later on, the constructive proposals were outperformed by meta-heuristic algorithms that introduced general purpose and easy-to-apply optimization paradigms.

Although meta-heuristics have become the main tool to adopt, contrary to constructive heuristics, they are evaluation-intensive algorithms, i.e., they need to exhaustively evaluate thousands or even millions of candidate solutions before arriving at a decision, making them impractical for scenarios with limited budget or online-streaming optimization \cite{chen2018foad}.

Algorithms based on Neural Networks (NN) play a crucial role in this regard. In recent years, Deep Learning (DL) techniques have exhibited remarkable performance across various machine learning tasks, drawing the attention of researchers from diverse domains, including optimization. As outlined in different reviews \cite{talbi2021machine, mazyavkina2021reinforcement, bengio2021machine}, DL-based approaches have been proposed as standalone solvers, parameter and/or operation selection methods, or as a component of more powerful hybrid algorithms.
Our focus in this work is on standalone (end-to-end) models, highlighting their capabilities and the avenues for further research aimed at enhancing their performance. Once trained, these models can rapidly make decisions, such as determining the next change required.
The first works in the topic proposed models, known as {\it constructive methods}, which generate a unique solution incrementally by iteratively adding an item to a partial solution until it is completed~\cite{bello2016neural, kool2018attention, kwon2020pomo}. Conversely, later papers have introduced \textit{improvement methods} that take a candidate solution and suggest a modification to improve it~\cite{chen2019learning, lu2019learning, wu2021learning}. In fact, this improvement process can be repeated iteratively, using the modified solution as the new input of the model. The reported results, although less competitive than state-of-the-art meta-heuristics for the most trending problems~\cite{accorsi2022guidelines, garmendia2022neural}, have captured the attention of the optimization research community as they were unimaginable some years ago. In fact, many of these proposals have outperformed the classical constructive heuristics.

However, looking at the progress of the research, we realize that the majority of the works have mainly illustrated their contributions on the Travelling Salesman Problem (TSP) \cite{applegate2011travelling} and other similar routing problems. Particularly, most models work on the idea that, when considering the graph representation of the COPs, the information is embedded node-wise~\cite{kool2018attention, kwon2020pomo}. However, there are problems such as Preference Ranking Problem (PRP) ~\cite{heckel2018approximate} or Graph Partitioning Problem (GPP) ~\cite{andreev2004balanced}, where the relevant information of the problem is edge-wise or even both node- and edge-wise. In these scenarios, node-wise proposals do not use all the available (and meaningful) information.

In line with the idea that future generation algorithms will come from the combination of meta-heuristic algorithms alongside machine learning models \cite{bengio2021machine}, we propose a new optimization framework which can replace and improve the traditional local-search-based methods by incorporating neural improvement models. Specifically, our contribution is twofold: (1) we present a neural improvement model to solve problems where the information is stored either in the nodes, in the edges, or in both of them, and (2) we show that the model can be used alone, or can be incorporated as a building block, for example for hill-climbing-based algorithms, to efficiently guide the selection of neighborhood operations.

In order to demonstrate the versatility and efficacy of the proposed framework, we conducted experiments across various optimization problems, including PRP, TSP, and GPP. The NI model, trained on node- and edge-features, demonstrated outstanding performance across all three problems, with exceptional results for PRP. It consistently recommended the best or near-best neighbors for each problem, and outperformed traditional methods in all cases.


The rest of the paper is organized as follows. Section \ref{related_work} introduces the most prominent works tackling the development of NN models for CO, both in a constructive and improvement manner. With illustrative purposes, we present the PRP in Section \ref{problem}, and propose a NI model in Section \ref{method}. A set of experiments is carried out in Section \ref{experiments} and the generalization of the model to other problems is discussed in Section \ref{transferability}. Finally, Section \ref{conclusion} concludes the paper.

\section{Related Work}
\label{related_work}

Although Neural Networks (NN) have been used since the decade of the 80s to solve COPs in the form of Hopfield Networks \cite{hopfield82}, it is only recently \cite{bengio2021machine, mazyavkina2021reinforcement} that advancements in computation power and the development of sophisticated architectures have enabled more efficient and increasingly competitive applications. As mentioned previously, NN-based optimization methods can be divided into two main groups according to their strategy.

\paragraph{Neural Constructive Methods} Most of the DL-based works develop policies to learn a constructive heuristic. These methods start from an empty solution and iteratively add an item to the solution until it is completed. In one of the earliest works in the \textit{Neural Combinatorial Optimization} paradigm, Bello \textit{et al.} \cite{bello2016neural} used a Pointer Network model \cite{vinyals2015pointer} to parameterize a policy that constructs a solution, item by item, for the TSP. Motivated by the results in \cite{bello2016neural}, and mainly focusing on the TSP, DL practitioners have successfully implemented different architectures such as Graph Neural Networks (GNN) \cite{cappart2021combinatorial, joshi2020learning} or Attention-based Networks \cite{kool2018attention, kwon2020pomo}.

Since the performance of the baseline models is still far from optimality (mostly in instances with more than a few hundred nodes), they are usually enhanced with supplementary algorithms, such as active search \cite{bello2016neural}, sampling \cite{kool2018attention} or beam search \cite{vinyals2015pointer}, which augment the solution diversity at the cost of increasing the computational time. As will be seen in the following, improvement methods offer a more efficient alternative, directly learning the transition from the current solution to a better one.

\paragraph{Neural Improvement Methods} Neural Improvement (NI) methods depart from a given solution, and iteratively propose a (set of) modification(s) to improve it until the solution cannot be further improved. NI methods utilize the learned policy to navigate intelligently across the different neighborhoods.

To that end, the architectures previously used for constructive methods have been reused for implementing improvement methods. Chen \textit{et al.} \cite{chen2019learning} use Long Short-Term Memory (LSTM) to parameterize two models: the first model outputs a score or probability for each region of the solution to be rewritten, while a second model selects the rule that modifies that region. Lu \textit{et al.} \cite{lu2019learning} use the Attention-based model to select a local operator among a pool of operators to solve the capacitated vehicle routing problem (VRP). Using also an attention network, Hottung \textit{et al.} \cite{hottung2019neural} propose a Neural Large Neighborhood Search that suggests new solutions destroying and repairing parts of the current solution.

Closer to our proposal, but limited to routing problems, Wu \textit{et al.} \cite{wu2021learning} train a policy that selects the node-pair to apply a local operator, e.g. 2-opt. Similarly, da Costa \textit{et al.} \cite{da2021learning} generalize the prior work to select k-opt operators. Falkner \textit{et al.} \cite{falkner2023learning} propose an improvement method to tackle the job scheduling problem which learns how to control the local search in three aspects: acceptance of the solution, neighborhood selection and perturbations. We have summarized the characteristics of the most relevant NI works in Table \ref{tab:literature}.

Improvement methods do not only incorporate the stationary instance data, but also need to consider the present solution. In fact, the difficulty of encoding the solution information into a latent space is a major challenge for most of the combinatorial problems.

As an example, there are various ways of representing solutions in routing problems. Each node (or city) can maintain a set of features that indicate the relative position in the current solution, such as the location and distance to the previously and subsequently visited nodes \cite{lu2019learning}. However, this technique does not consider the whole solution as one, as it only contemplates consecutive pairs of nodes in the solution. A common strategy in this case is to incorporate Positional Encodings (PE), which capture the sequence of the visited cities (nodes) in a given solution \cite{kool2018attention}. Recently, Ma \textit{et al.} \cite{ma2021learning} proposed a \textit{cyclic PE} that captures the circularity and symmetry of the routing problem, making it more suitable for representing solutions than the conventional PE. 

Nevertheless, in some graph problems the essential information is codified solely in the edges, and thus, prior methods that focus on node embeddings \cite{lu2019learning, ma2021learning} are not capable of properly encoding the relative information.

Even though there are few works that consider edge weights to encode problem specific features \cite{joshi2019efficient, fu2021generalize}, they focus on creating a heatmap of probabilities for each edge to belong to the optimal solution, and use it to construct (or sample) a (set of) solution(s).
In this work, similar to \cite{joshi2019efficient, fu2021generalize}, we utilize nodes and edges to represent graph data. However, what sets our approach apart is its ability to encode both instance and solution information, and then use this encoded information to propose a local improving move.
In addition, when compared to \cite{wu2021learning}, we provide some guidelines for generalizing to different graph-based problems, by considering both node and edge features. Furthermore, unlike \cite{wu2021learning}, we do not rely on positional encodings to embed current solution information, as we naturally embed it in the edge features. Lastly, we propose a standalone neural improvement model and demonstrate how to combine it with classical local search techniques like multi-start hill climbing, tabu search, or iterative local search.

In the following section, we will present an optimization problem that illustrates the need to develop new NI models that also consider edge features.

\begin{table*}
  \caption{Analysis of the most relevant Neural Improvement works in the literature.}
  \label{tab:literature}%
    \footnotesize
  \centering

\begin{tabular}{*{5}{p{.03\linewidth} p{.15\linewidth} p{.1\linewidth} p{.25\linewidth} p{.3\linewidth}}}
\toprule
Ref & Problems  & Architecture & Dynamic input & Functionality and output  \\ \midrule
\cite{chen2019learning}  & ES, JS, VRP & LSTM &  Prior and posterior node information & Region picking and Rule selector modules  \\ 
\cite{lu2019learning}  & CVRP & Attn Net & Prior and posterior node information & Operator selection from a pool of operators  \\ 
\cite{da2021learning}  & TSP &  GCN, RNN & Recurrent Neural Networks & Select where to apply k-opt  \\ 
\cite{hottung2019neural}  & CVRP & Attn Net & Only considers parts of the solution & Large Neighborhood Search: repair and destroy operators  \\ 
\cite{falkner2023learning}  & JS, CVRP & GNN & Boolean edge features & A model that accepts local search moves and selects the operator and when to perturb  \\ 
\cite{wu2021learning}  & TSP, CVRP & Transformer & Positional Encoding & Node-pair modification \\ 
\cite{ma2021learning}  & TSP, VRP, CVRP & Transformer & Cyclic Positional Encoding & Node-pair modification  \\ 
[0.1cm]
\bottomrule
\end{tabular}
\end{table*}%

\section{Preference Ranking Problem}
\label{problem}

Ranking items based on preferences or opinions is, in general, a straightforward task if the number of alternatives to rank is relatively small. Nevertheless, as the number of alternatives/items increases, it becomes harder to get full rankings that are consistent with the pairwise item-preferences. Think of ranking 50 players in a tournament using their paired comparisons from the best performing player to the worst. Obtaining the ranking that agrees with most of the pairwise comparisons is not trivial. This task is known as the Preference Ranking Problem (PRP) \cite{heckel2018approximate}. Formally, given a preference matrix $B=[b_{ij}]_{N \times N}$ where entries of the matrix $b_{ij}$ represent the preference of item $i$ against item $j$, the aim is to find the simultaneous permutation $\omega$ of rows and columns of $B$ so that the sum of entries in the upper-triangle of the matrix is maximized (see Eq. \ref{eq:lop_of}).

\begin{figure}
    \centering
    \includegraphics[width=0.47\textwidth]{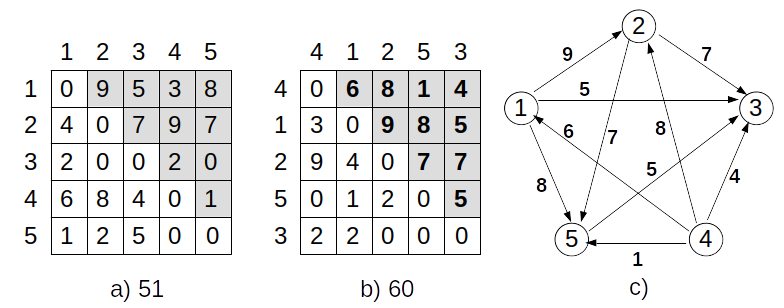}
    \caption{Example of a PRP instance of size $N=5$. \textbf{a}) A preference matrix of size $N=5$ ordered as the identity permutation $\mathbf{\omega}_e = ($1 2 3 4 5$)$. Entries of the matrix contributing to the objective function are highlighted in grey, the sum of the entries in the upper diagonal gives the objective value, which is 51. \textbf{b}) For this instance, the optimal solution is given by the permutation $\mathbf{\omega}_{opt} =  ($4 1 2 5 3$)$ with an objective value of 60. \textbf{c}) Equivalent graph representation of the optimal solution. Edge weights denote the preference given by the optimal solution.}
    \label{fig:prp}
\end{figure}

Note that row $i$ in $B$ describes the preference vector of item $i$ over the rest of $N-1$ items, while column $i$ denotes the preference of the rest of the items over item $i$. Thus, in order to maximize the upper triangle of the matrix, preferred items must precede in the ranking (see Fig.\ref{fig:prp}a).

\begin{equation}
\label{eq:lop_of}
f(\mathbf{\omega}) = \sum_{i=1}^{N-1} \sum_{j=i+1}^{N} b_{\mathbf{\omega}(i) \mathbf{\omega}(j)}
\end{equation}

In addition to the matrix representation (Fig. \ref{fig:prp}a), the problem can be formulated as a complete bidirected graph where nodes represent the set of items to be ranked and the weighted edges denote the preference between items. A pair of nodes $i$ and $j$ has two connecting edges $(i,j)$ and $(j,i)$, with weights $b_{ij}$ and $b_{ji}$ that form the previously mentioned preference matrix $B$. A solution (permutation) to the PRP can be also represented as an acyclic tournament on the graph, where the node (item) ranked first has only outgoing edges, the second in the ranking has 1 incoming edge and the rest are outgoing, and so on until the last ranked node, which only has incoming edges (see Fig.\ref{fig:prp}b and c).

Ranking from pairwise comparisons is a ubiquitous problem in modern Machine Learning research. It has attracted the attention of the community due to its applicability in various research areas, including, without being limited to: machine translation \cite{tromble2009learning}, economics \cite{leontief1986input}, corruption perception \cite{achatz2006corruption} or any other task requiring a ranking of items, such as sport tournaments, web search, resource allocation and cybersecurity \cite{anderson2019rankability, cameron2021linear, shah2013case}.

\section{Method}
\label{method}

The idea of solving a graph problem iteratively with a NI model can be formulated as a Markov Decision Process (MDP), where a policy $\pi$ is responsible for selecting an action $a$ at each step $t$ based on a given state $s_t$ of the problem. The main entities of the MDP in this work can be described as:

\begin{itemize}
    \item \textbf{State.} A state $s_t$ represents the information of the environment at step $t$. In this case, the state gathers data from two information sources: (1) stationary data, i.e., the instance to be solved; and (2) dynamic data, that is, the current solution $\omega_t$ for the problem at step $t$. 
    \item \textbf{Action.} At every step, the learnt policy selects an action $a_t$, which involves a pair of items of the current solution that, according to the policy, need to be modified. Once selected, an operator will be applied, modifying the current solution. Note that either one of the items, both, or more may be involved in the modification, depending on the operator.
    \item \textbf{Reward.} The transition between states $s_t$ and $s_{t+1}$ is derived from an operator applied to a pair of items given by $a_t$. The reward function represents the improvement of the solution quality across states. Different function designs can be used, as will be explained in Section \ref{learning}.
    
\end{itemize}

In what follows, a detailed description of the NI model is provided. Even though the design is general for any graph-based problem, for the sake of clarity, we provide illustrative examples based on the PRP. Extensions to other problems will be discussed later in Section \ref{transferability}.

\subsection{Neural Improvement Model}
\begin{figure*}
    \centering    
    \includegraphics[width=0.88\textwidth]{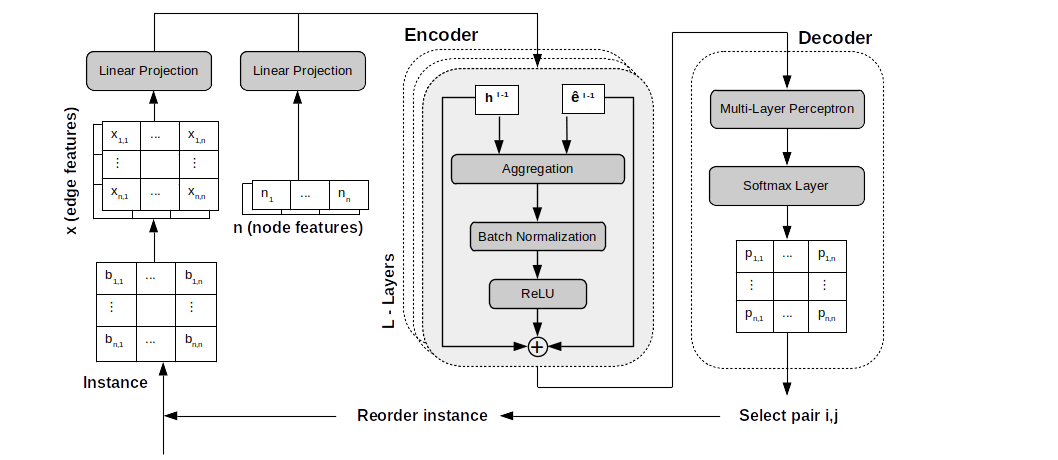}
    \caption{High level design of the NI model architecture. Node and edge features are linearly projected and fed to the encoder (GNN). Edge embeddings computed through $L$ NN layers are passed to the decoder (MLP) and this outputs the edge probabilities, which will be later used to select the pair of nodes in which an operator is applied.}
    \label{fig:architecture}
\end{figure*}

We will parameterize the policy $\pi$ as a NN model with trainable parameters $\theta$.
Considering the case study presented in Section \ref{problem} the model architecture need to meet some requirements: (1) it needs to codify graph structure data, (2) it needs to be invariant to input permutations, (3) invariant to input size changes, and (4) it needs to consider the solution information.
Considering that, we opted to use a GNN encoder, capable of gathering both node and edge features, and a Multi-Layer Perceptron (MLP) decoder, a simple and fast architecture that interprets the embedded features and generates a probability distribution over the set of possible actions. Fig. \ref{fig:architecture} presents the general architecture of the model.
Apart from the presented model, we have analyzed two different encoder and decoder architectures and tested various hyperparameters. The outcomes of these experiments are presented in Appendix \ref{app1}.

\subsubsection{Encoder} 
Given a fully connected graph with $N$ nodes, there are $N \times N$ edges or node pairs, and each edge $(i,j)$ has a weight $b_{ij}$ that represents the relative information of node $i$ with respect to node $j$. Note that only $N \times (N-1)$ edges need to be considered since edge $(i,i)$ does not provide any useful information. As previously noted, the policy considers both instance information (stationary) and a candidate solution at time step $t$ (dynamic). For this purpose, we will use a bi-dimensional feature vector $\mathbf{x}_{ij} \in \mathbb{R}^{2}$ for each edge $(i,j)$. The first dimension represents whether node $i$ precedes node $j$ in the solution. If this is the case, it is set to $b_{ij}$ (edge weight) or it is set to zero otherwise. Similarly, the second dimension denotes the opposite, that is, whether node $j$ precedes node $i$.


For the PRP, nodes do not reflect any problem-specific information, and thus all the nodes are initialized with the same value. In fact, following a similar strategy to that proposed by Kwon \textit{et al.} \cite{kwon2021matrix}, we use the identity vector as node features, $\mathbf{n} \in \mathbb{R}^{N}$. Even though all the nodes are initiated with the same value, their participation is required, as they spread edge features across the graph in the encoding: node $i$ will gather information from edges $(i, k)$ and $(k, i)$,  $k = 1, ..., N$.

Node and edge features will be linearly projected to produce \textit{d}-dimensional node- $\mathbf{h}_{i}  \in \mathbb{R}^{1 \times d}$ and edge- $\mathbf{e}_{ij}  \in \mathbb{R}^{1 \times d}$ embeddings
\begin{equation}
\label{eq:node_embeddings}
\mathbf{h}_{i} = n_{i} * V_h + U_h
\end{equation}
\begin{equation}
\label{eq:edge_embeddings}
\mathbf{e}_{ij} = \mathbf{x}_{ij} * V_e + U_e
\end{equation}
where $V_e$ $\in \mathbb{R}^{2 \times d}$ and $V_h$, $U_e$ and $U_h$ $\in \mathbb{R}^{1 \times d}$ are learnable parameters.

The encoding process consists of $L$ Graph Neural Network (GNN) layers (denoted by the superscript $l$) that perform a sequential message passing between nodes and their connecting edges. This enables the GNN layers to learn rich representations of the graph structure and capture complex relationships, such as precedences of items in the solution (see left part of Fig. \ref{fig:architecture}). Eqs. \ref{eq:node} and \ref{eq:edge} define the message passing in each layer, where $W_1^l$, $W_2^l$, $W_3^l$, $W_4^l$ and $W_5^l$ $\in \mathbb{R}^{d \times d}$ are also learnable parameters, $BN$ denotes the batch normalization layer, $\sigma$ is the sigmoid function and $\odot$ is the Hadamard product.

\begin{equation}
\label{eq:node}
  \mathbf{h}_{i}^{l+1} = \mathbf{h}_{i}^{l} + ReLU\left(BN\left(W_1^l \mathbf{h}_i^l + \sum_{j=1}^N (\sigma(\mathbf{e}^l_{ij}) \odot W_2^l \mathbf{h}_j^l)\right)\right)
\end{equation}
\begin{equation}
\label{eq:edge}
  \mathbf{e}_{ij}^{l+1} = \mathbf{e}_{ij}^{l} + ReLU\left(BN\left(W_3^l \mathbf{e}_{ij}^l + W_4^l \mathbf{h}_{i}^l + W_5^l \mathbf{h}_{j}^l\right)\right)
\end{equation}

The output of the encoder, which is fed to the decoder, consists of the edge embeddings of the last layer $\mathbf{e}_{ij}^L$.

\subsubsection{Decoder}
Edge embeddings are fed to the decoder, a MLP that converts the edge embeddings into logits $\mathbf{u_{ij}}$ in a format that can be used to select the next operator;
\begin{equation}
 \label{eq:logits}
 \mathbf{u_{ij}} = \left\{
       \begin{array}{ll}
	  \mathrm{MLP}(e_{ij}) & \mathrm{if\ } i \neq j\;\;  \\
	 - \infty & \mathrm{otherwise} \\
       \end{array}
     \right.
\end{equation}

The logits are then normalized using the \textit{Softmax} function to produce a matrix $\mathbf{p} \in \mathbb{R}^{N \times N}$ which gives the probability of modifying the pair of items ($i$,$j$) in the candidate solution. 

\subsection{Learning}
\label{learning}

The improvement policy will be learned using the REINFORCE algorithm \cite{williams1992simple}. Given a state $s_t = (B, \mathbf{\omega}_t)$ which includes an instance $B$ and a candidate solution $\mathbf{\omega}_t$ at step $t$, the model gives a probability distribution $p_\theta (a_t|s_t)$ for all the possible pairs of items to be modified. After performing an operation $O(\mathbf{\omega}_t|a_t)$ with the selected pair, a new solution $\mathbf{\omega}_{t+1}$ is obtained. The training is performed minimizing the following loss function
\begin{equation}
\label{eq:loss}
 \mathcal{L}(\theta | s) = \mathbb{E}_{p_\theta (s, \mathbf{\omega}_t)} [-R_t \log p_\theta (s, \mathbf{\omega}_t)]
\end{equation}
by gradient descent, where $R_t = \sum\limits_{i=0}^{T-1} \gamma^{i}(r_{t+i} - r_{t+i-1})$ corresponds to the sum of cumulative rewards $r_i$ with a decay factor $\gamma$ in an episode of length $T$.%

This artefact is a key piece of the model. It has been conceived to avoid myopic behaviors, permitting short-term and long-term strategies, as a different number of operations are allowed ($T$) before evaluating the quality of the movement sequence. In addition, the decay factor offers the practitioner a way to weight every movement, paying, for example, more attention to the earliest movements.


\subsubsection{Reward functions}
Different Reward Functions (RF) have been proposed in the literature for obtaining $r$. Lu \textit{et al.} \cite{lu2019learning} use a reward function (RF1) that takes the objective value of the initial solution as the baseline and, for each subsequent action, the reward at step $t$ is defined as the difference between $f(\mathbf{\omega}_t)$ and the baseline. The drawback of this function is that rewards may get larger and larger, and moves that worsen the sequence can be given positive reward (as they are better than the baseline).

\begin{figure}
\centering
\includegraphics[width=0.47\textwidth]{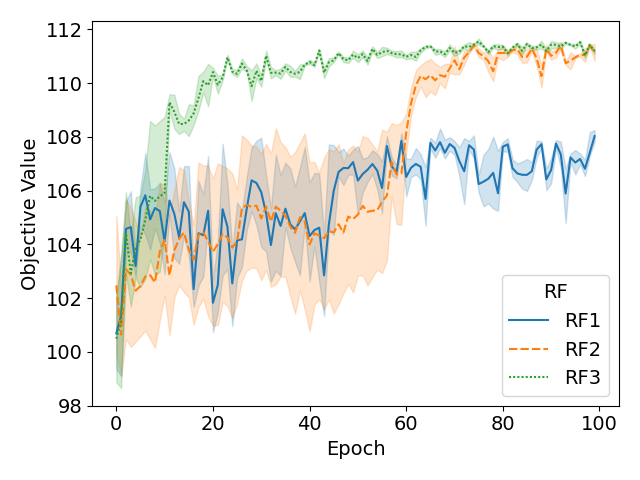}
\caption{Comparison of the three reward functions RF1, RF2 and RF3. Model has been trained for 100 epochs, and at each epoch a model checkpoint is saved and used to solve a test benchmark of 512 randomly generated PRP instances of size 20. Obtained objective values are saved to form the curves.}
\label{wrap-fig:1}
\end{figure} 

Alternatively, the most common approach in recent works \cite{wu2021learning, ma2021learning}, is to define the reward (RF2) as $r_t = \max [f(\mathbf{\omega}_{t+1}), f(\mathbf{\omega}^*_t)] - f(\mathbf{\omega}^*_t)$, where $f(\mathbf{\omega}^*_t)$ is the objective value of the best solution found until time $t$. Note that this alternative yields only non-negative rewards, and all the actions that do not improve the solution receive an equal reward $r_t = 0$. In our case, we propose a simple but effective reward function (RF3) $r_t = f(\mathbf{\omega}_{t+1}) - f(\mathbf{\omega}_t)$, which defines the reward as the improvement of the objective value between steps $t$ and $t+1$, and also considers negative values. RF3 reward function yields a faster convergence with less variability, as can be seen in the comparison of the mentioned reward functions depicted in Fig. \ref{wrap-fig:1}.

\subsubsection{Automated Curriculum Learning} 
Curriculum Learning is a training strategy that involves controlling the difficulty of the samples throughout the process, where the difficulty level is manually increased \cite{bengio2009curriculum}. In this context, difficulty is measured by calculating the percentage of moves that worsens the objective value in relation to all the possible moves.
However, in our approach, we do not utilize a manual curriculum learning strategy like the one employed by Ma et al. \cite{ma2021learning}. Instead, we employ an iterative approach where the model receives the solution that has been modified in the previous iteration. As the model improves and suggests better moves for the current instances during training, the number of improving moves decreases, leading naturally to increase the difficulty level. This iterative approach enables an automated curriculum learning, which is valid for prior optimization problems and does not require neither any problem knowledge nor external intervention.

However, learning is performed with a batch of instances and not all of them reach a local optimum in the same number of steps. Thus, we save the best average reward obtained by the model, and we consider the algorithm to be stuck when it does not improve the best average reward for $K_{max}$ iterations. 
A large value of $K_{max}$ should give the model more chances to visit higher quality solutions, but can also introduce undesired computational overheads.

\subsubsection{Operator} 
\begin{figure}
    \centering
    \includegraphics[width=0.47\textwidth]{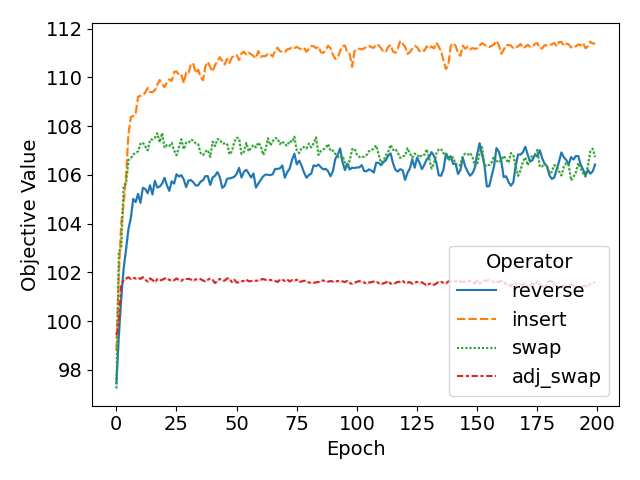}
    \caption{Training curves using different pairwise operators. Each operator is used in a training procedure of 200 epochs executed 5 times with a different random seed. The model output is masked so that, for each operator, only feasible pairs can be selected.}
    \label{fig:operator}
\end{figure}

The model is flexible, allowing the practitioner to define the operator that best fits the problem at hand. In the particular case of the PRP, there are several operators that could be applied in order to perform the pairwise modification, such as, \textit{insert}, \textit{swap}, \textit{adjacent-swap} and \textit{reverse} operators\footnote{Given an edge $(i,j)$, denoting the items in positions $i$ and $j$ in the solution: the \textbf{insert} operator consists of removing the item at the position $i$ and placing it at position $j$, the \textbf{swap} exchanges the items at both positions, the \textbf{adjacent-swap} is a swap that only considers adjacent positions, and the \textbf{reverse} operator reverses the sub-permutation between positions $i$ and $j$.}. Previous works \cite{laguna1999intensification} have demonstrated that the \textit{insert} operator yields the best results for the PRP. In fact, we confirm it in Fig. \ref{fig:operator}, where we show the comparison of the training convergence curves for the mentioned pairwise operators.

\subsubsection{Training algorithm} The implemented training algorithm is presented in Algorithm \ref{alg:ni}. The state $s_t$ represents the instance and a candidate solution at time step $t$. At each epoch a random state is generated with a random solution (line 3). Then, the model gives the probability vector (line 7), and an action is sampled and applied to the current solution (lines 8-9). The best found reward is saved (line 10) and the process is repeated until the model does not improve the best known solution for a specified number of steps ($K_{max}$) consecutively. For the sake of clarity, Algorithm \ref{alg:ni} shows the training with one instance, even though a batch is used. In order to use a batch of instances, states $\mathbf{s}_t$, probabilities $\mathbf{p}_{\theta}$, actions $\mathbf{a}_t$ and rewards $\mathbf{r}_t$ will be vectors and the average of $\mathbf{r}_t$ will be used to control the improvement condition (line 11).

\begin{algorithm}
\caption{Training Algorithm for the NI Model}
\label{alg:ni}
\begin{algorithmic}[1]
\Require total number of epochs $n_{epochs}$, stopping criterion $K_{max}$, operator $O$, episode length $T$, learning rate $\alpha$ and discount factor $\gamma$.
\State Initialize the policy $\pi_{\theta}$ with random weights $\theta$ 
\For{$epoch = \{1, ..., n_{epochs}\}$}
\State State $s_0 \gets \textsc{RndmGen.}$, step $t \gets 0$, count $k \gets 0$ 
\State $r_{best} \gets$ \textsc{ComputeReward}($s_0$)

\While{$k \leq K_{max}$}
\State $t \gets t+1$
\State $p_{\theta} \gets \pi_{\theta}(s_{t-1})$
\State $a_t \gets$ \textsc{SampleAction}($p_{\theta}$) 
\State $s_t \gets O(s_{t-1}, a_t)$
\State $r_t \gets$ \textsc{ComputeReward}($s_t$)
\State $(r_{best},k) \gets$ \textsc{UpdateBest}($r_t,r_{best},k$)

\If{$t \bmod T = 0$}
\State $R = \frac{1}{T} \sum\limits_{i=t-T}^t( \sum\limits_{j=i}^t \gamma^{j-i}(r_i - r_{i-1}))$
\State $\nabla_{\theta}J(\theta) \gets -R  \nabla_{\theta} \log p_{\theta}(a_t)$
\State $\theta \gets \theta + \textsc{GradientClip}(\alpha\nabla_{\theta}J(\theta)$)
\EndIf
\EndWhile
\EndFor
\end{algorithmic}
\end{algorithm}

\subsection{Applications of the Neural Improvement model}
\label{applications}

The Hill Climbing heuristic (HC) is a procedure that continuously tries to improve a given solution performing local changes (for example swapping two items) and looking for better candidate solutions in the neighborhood. Examples of conventional HC procedures include, among others, \textit{Best First Hill Climbing} (BFHC), which selects the first found candidate (neighbor) that improves the present solution; \textit{Steepest-Ascent Hill Climbing} (SAHC), which exhaustively searches all the neighborhood and selects the best candidate solution from it; or \textit{Stochastic Hill Climbing} (SHC), which randomly picks one solution from the neighborhood.

\subsubsection{Neural Hill Climber}
With the goal of reducing the large number of evaluations needed by conventional HC, we propose the Neural Hill Climber (NHC), which attempts to suggest the best neighbor based on a given solution and then uses this neighbor as the starting point for the next iteration. This cycle is repeated until the best possible solution is found. In general, HC heuristics do not allow the objective value to decrease. In our approach, to ensure that the selected action is an improving move, we sort the probability vector given by the model and select the first action that improves the solution. Of course, other strategies could be also adopted.

\subsubsection{Advanced Hill Climbers}
Eventually, just as conventional HC procedures, the NHC will get stuck in a local optimum where an improving move cannot be found. More advanced algorithms have been proposed in the literature that try to escape the local optima by performing a restart or a perturbation to the current solution \cite{blum2003metaheuristics}.
One of the many examples is the Multi-Start HC (msHC), which restarts the search departing from another random candidate solution whenever an improving move cannot be found.

An alternative to msHC is the Iterated Local Search (ILS) \cite{lourenco2019iterated}. Once the search gets stuck in a local optimum, instead of restarting the algorithm, ILS perturbs the best solution found so far and the search is resumed in this new solution. The perturbation level is dynamically changed based on the total budget left (number of evaluations or time). 

The third example considered in this paper is the Tabu Search (TS) \cite{glover1993user}, which enhances the performance of the HC method allowing worsening moves whenever a local optimum is reached. In order to avoid getting trapped in cycles, TS maintains a tabu memory of previously visited states to prevent visiting them again in the next $m$ moves. 

We will use the NI model to guide the local moves of a Multi-Start Neural Hill Climbing (msNHC), a Neural Iterated Local Search (NILS) and a Neural Tabu Search (NTS), and analyse their performance compared to conventional versions in the following section\footnote{For TS, a short-term memory of size 200 is used. For ILS, once a local optima is reached, the solution is perturbed with $n$ random swaps, $n$ being calculated based on the remaining evaluations ($RE$) in the execution: $n = \lfloor(N/2) * (RE/E)\rfloor$, where $N$ is the problem size and $E$ denotes the evaluation budget.}.

\section{Experiments}
\label{experiments}

In this section, we present a thorough experimentation of the proposed NI model. First, we analyse the performance of the NI model in the short-term (\textit{one-step}) and long-term (\textit{multi-step}) basis. Afterwards, we test the efficiency of the NI model implemented as a building block of HC algorithms.

\subsection{Setup} 
\label{setup}

For the experiments, we deploy a 3-layer GNN as encoder, being the embedding dimension 128. As the decoder, we use a MLP of 4 layers with hidden dimensions (128, 64, 32 and 1) and ReLU activation. 
Regarding the training hyperparameters, a learning rate is set to 1e-4, the episode length to $T=20$, the decay factor to $\gamma = 0.1$ and the maximum number of consecutive non-improving moves to $K_{max}=5$. Further details on the selection of the hyperparameters can be found in Appendix \ref{app1}.
To train the NI model, we adopt a common practice of utilizing randomly generated instances. Each epoch, a different batch of 64 instances is used, and gradients are averaged across this batch to update the model parameters. Due to limited computational resources, two different models have been trained using sizes $N=20$ and $N=40$ respectively. Both models have been trained for 5000 epochs\footnote{Even though we use instances of the same size to train each model, note that, due to the element-wise operations performed in the encoder, it is possible to combine instances of different sizes.}.

Related to the size of the instances used for training DL-based models, a common drawback is the lack of scalability: most of the DL-based models need to be trained with instances of the same size to those used later for inference. However, the introduced model can be trained with a computationally affordable instance size, and then be used for solving larger size instances (discussed later in this section). If not mentioned differently, for large sizes, the model trained with instances of size ($N=20$) will be used.

We have implemented the algorithms using \textit{Python 3.8}. Neural models have been trained in an \textit{Nvidia RTX A5000} GPU, while methods that do not need a GPU are run on computers with \textit{Intel Xeon X5650} CPUs and 64GB of memory.

\subsection{NI Model Performance Analysis} 

During the optimization process, obtaining solutions that improve the current one becomes harder at each step. With the aim of testing the ability of the model, we present two scenarios: (1) One-Step, that is, departing from a random solution, is the model able to find the best neighbor?, and (2) Multi-Step, departing from a random solution, is the model able to propose a better solution repeatedly, improving the previously found solution in a consistent way throughout the optimization process?

\subsubsection{One-step} 
\begin{figure}
    \centering
    \includegraphics[width=0.47\textwidth]{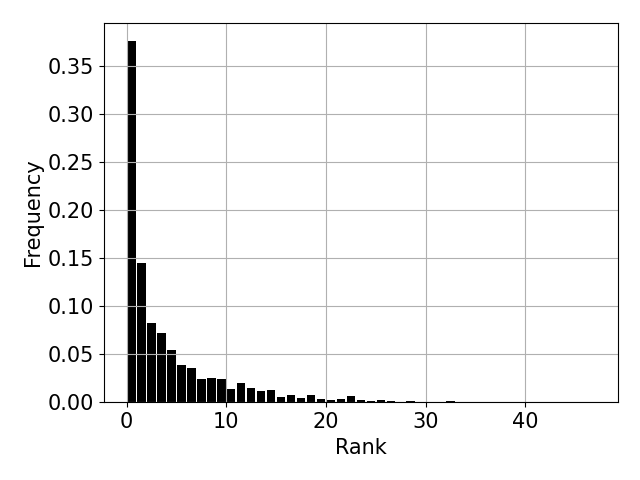}
    \caption{Histogram that shows the ranking of the action selected by the model among all the possible actions on 2000 different instances of size $N=20$. The first rank denotes the best possible action (based on the one-step improvement). Note that the x-axis is cut since none of the proposed actions have been ranked between the 50th and 361st positions.}
    \label{fig:rank_hist}
\end{figure}

We focus on the capability of the NI model to provide a solution that outperforms the present one. In terms of neighborhood, we expect the NI model to be able to identify the best or at least one of the best neighboring moves.
We perform One-step predictions for 2000 instances of size $N=20$. Results are depicted in Fig. \ref{fig:rank_hist} in the form of a histogram. Specifically, we calculated the ranking of the predicted move among all the available actions ($(N-1)^2$, 361).
Note that the model selects the best neighbor in more than 35\% of the times. Considering all the possible insert operations, on average, the action that the model takes is in the $99^{th}$ percentile, ranked $4^{th}$ out of $361$.

In larger instance sizes, even if trained with $N=20$, the NI model is not only able to maintain the good performance, but also improves the percentile rank of the selected action. It selects, on average, the $13^{rd}$ best action out of $2401$ for $N=50$ and the $32^{nd}$ out of $9801$ for $N=100$.

\subsubsection{Multi-step}
\label{long_term_exp}
\begin{figure}
    \centering
    \includegraphics[width=0.46\textwidth]{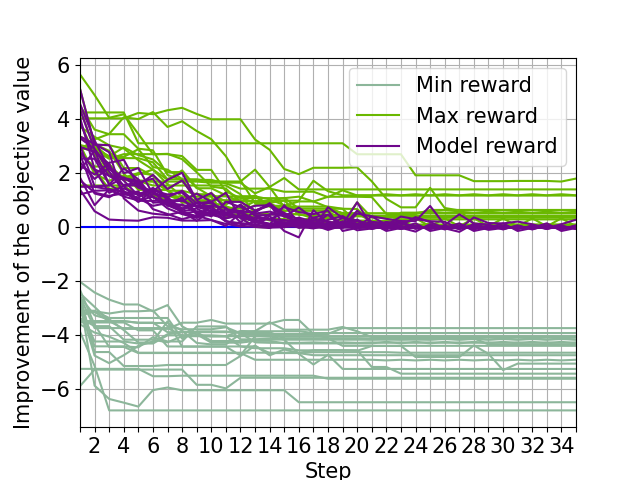}
    \caption{Line plot of 20 different executions of the NHC. Each purple line represents the reward, or the improvement of the objective value (y-axis), given by the selected action over consecutive steps (x-axis). The minimum and maximum rewards that could be obtained in each step are also shown. Instances of $N=20$ are solved using models trained with randomly generated instances of the same size.}
    \label{fig:parallel}
\end{figure}

Once tested that the NI model successfully learns to select one of the best possible actions (one-step), we still need to verify whether it obtains increasingly better solutions in a multi-step scheme. For that purpose, we let the model make consecutive moves, being fed, at each iteration, with the solution obtained in the previous one. Fig. \ref{fig:parallel} illustrates the behaviour of the model for consecutive steps. The figure shows that: (1) as expected the maximum obtainable reward decreases over the improvement steps, presumably increasing the difficulty, and (2) the action selected by the model is closer to the maximum reward than the minimum reward, which confirms the good performance of the model when several steps are completed.

%
\subsection{Neural Hill Climbing Performance Analysis}

In this section we compare the performance of NHC to two conventional approaches: Steepest Ascent (SAHC) and Best First (BFHC). We let the algorithms run until they get trapped in a local optima, and repeat the optimization for 2000 different instances.
During the optimization run we computed the gap (\%) to the optimum objective value of the instance and the consumed evaluations.

\begin{figure}
    \centering
    \includegraphics[width=0.46\textwidth]{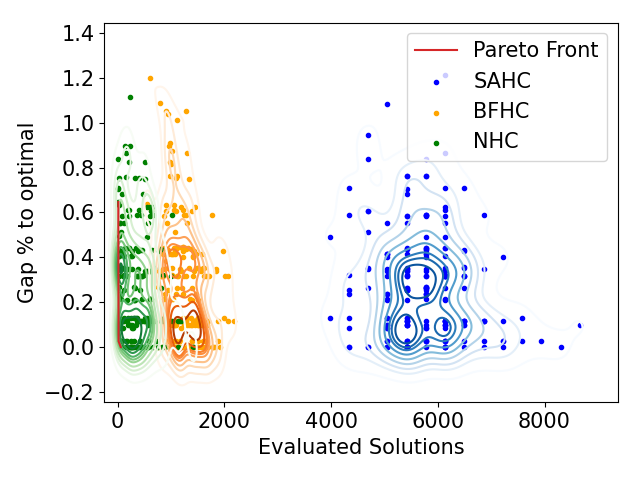}
    \caption{Bi-criteria analysis regarding the gap (\%) to optimal value and the total number of objective value evaluations. The lower-left corner is best. Instances of $N=20$ are solved using models trained with randomly generated instances of the same size.}
    \label{fig:mo_heatmap}
\end{figure}

Fig. \ref{fig:mo_heatmap} shows the results of the experiment. NHC, SAHC and BFHC obtain an average gap of 0.28\%, 0.29\% and 0.28\% respectively. Regarding the number of solution evaluations, NHC is the cheapest among all three procedures evaluating on average 355 solutions. 948 solutions are explored by BFHC, and 6293 by SAHC, being the most expensive procedure since it needs to evaluate all the possible actions (entire neighborhood) before performing a move. All the solutions in the pareto front belong to NHC (see the red line). In summary, NHC performs as well as SAHC and BFHC, but requires a significantly lower number of evaluations.

\subsection{Advanced Hill Climbers Performance Analysis} 

We expand here the analysis of the advanced algorithms explained in Section \ref{applications}. The performance is measured for several instance sizes (20, 50, 100 and 500) setting three different maximum number of evaluations: 10N, 100N and 1000N, being N the size of the instance. We incorporate the following algorithms for the comparison: msHC with conventional search strategies, such as Best First (\textbf{msBFHC}), Steepest Ascent (\textbf{msSAHC}) and Stochastic (\textbf{msSHC}); the neural version, guided with a NI model trained using instances of 20 and 40 (\textbf{msNHC-20}, \textbf{msNHC-40}); Tabu Search algorithm with an underlying best-first strategy (\textbf{BFTS}), the Neural Tabu Search \textbf{(NTS)} algorithm, an Iterated Local Search algorithm with a best-first selection (\textbf{BFILS}) and its neural version (\textbf{NILS}). As a baseline method, we also add the Becker constructive method \cite{Becker1967}. The performance is measured by means of the average gap percentage to the best known objective value, given by the state-of-the-art metaheuristic \cite{lugo2021diversity}. For each size, we use 512 randomly generated instances.

The results are presented in Table \ref{tab:performance}. As we can see, the neural variants (msNHC, NTS and NILS) outperform the respective conventional ones in all the cases. While conventional methods perform similarly, the neural multi-start method shows the largest improvement compared to its conventional variant.


\begin{table*}
\centering
  \caption{Comparison of improvement methods. Average gap (\%) to the best known value for different maximum number of solution evaluations (E). Best results amongst each kind of algorithms are highlighted in bold.}
  \label{tab:performance}%
  \centering
\bgroup
\def\arraystretch{1.2}
  \begin{tabular}{l|rrr|rrr|rrr|rrr}
 & \multicolumn{3}{c|}{N=20} & \multicolumn{3}{c|}{N=50}& \multicolumn{3}{c}{N=100} & \multicolumn{3}{c}{N=500}\\
 \hline
\textbf{Method} & E20 & E200 & E2000 & E50  & E500 & E5000 & E100  & E1000 & E10000 & E500  & E5000 & E50000 \\ 
\hline
Becker  &  & 3.38\% &  &  & 3.30\% &  &  & 2.87\% &   &   &  1.63\% &  \\
\hline
msBFHC   & 10.97\% & 2.58\%  & 0.28\% & 8.60\%  & 2.89\% & 0.68\% & 6.80\% & 2.67\% & 0.82\% & 4.03\% & 1.87\% &  0.73\%\\
msSAHC   & 13.59\% & 7.72\%  & 0.49\% & 10.92\% & 8.93\% & 2.35\% & 8.67\% & 7.91\% & 3.95\% & 4.51\% & 4.44\% &  3.89\%\\
msSHC    & 13.30\% & 10.21\% & 8.07\% & 10.60\% & 9.32\% & 8.42\% & 8.50\% & 7.86\% & 7.36\% &  4.44\%  & 4.31\%  &  4.22\% \\ 
msNHC-20 & 2.46\%  & \textbf{0.85\%}  & \textbf{0.14\%} & 2.93\%  & 1.34\%  & 0.65\%  & 3.42\% & 1.66\% & 0.79\% & 3.07\% & 1.71\% &  0.78\% \\
msNHC-40 &  \textbf{2.09\%} & 0.92\% & 0.16\% & \textbf{2.25\%} & \textbf{1.24\%} & \textbf{0.63\%} &  \textbf{2.98\%} &  \textbf{1.66\%} &  0.77\% & \textbf{2.92\%} & \textbf{1.59\%} &  \textbf{0.69\%} \\
\hline
BFTS    & 10.94\% & 2.56\%  & 0.36\% & 8.57\% & 2.88\% & 0.71\% & 6.79\% & 2.64\% & 0.81\% & 3.52\% & 1.72\% & 0.75\%  \\
NTS  & 2.46\%  & \textbf{0.85\%}  & 0.25\% & 2.93\%  & 1.34\%  & 0.64\%  & 3.42\%  & 1.67\% & \textbf{0.75\%} & 3.07\% & 1.69\% & 0.72\% \\
\hline
BFILS   & 10.98\% & 2.46\%  & 0.28\% & 8.55\% & 2.91\% & 0.68\% & 6.79\% & 2.65\% & 0.82\% & 3.52\% & 1.74\% & 0.74\%  \\
NILS & 2.46\%  & \textbf{0.85\%}  & 0.15\% & 2.93\% & 1.34\% & 0.63\% & 3.42\% & 1.66\% & 0.78\% & 3.07\%& 1.61\% & 0.71\% \\
\end{tabular}%
\egroup
\end{table*}%

Training this kind of DL-based models for large size instances becomes computationally demanding. Thus, designing a model that can be trained on small sizes and later applied to larger instances becomes advisable or, depending on the context, even mandatory. As shown in Table \ref{tab:performance}, models trained for sizes $20$ and $40$ are able to outperform their traditional counterpart on larger instances (up to 500). Particularly, even though msNHC-40 shows a better behavior compared to msNHC-20, the differences between them are quite small, demonstrating the proper generalization of the model in larger sizes. We further evaluate the generalization of the model trained with randomly generated instances to other types of instances in Appendix \ref{app2}.

\subsection{Computational time}
A common discussion in the optimization area is whether the number of evaluations or the computational time should be used when comparing different algorithms. However, it is important to note that different programming languages and hardware platforms may produce varying results. Therefore, this section has been added to supplement the previous one, by offering information on the required computational times for both training and inference of the presented model.

\subsubsection{Inference time}
The performance of the HC algorithms have been measured using a maximum number of solution evaluations as a limit. However, one could argue that neural methods have, for an equal number of evaluations, a higher time-cost compared to conventional ones; i.e., getting the next movement from the neural network is more costly than choosing a movement randomly or selecting the next movement of a greedy sequence.  Nevertheless, the reduced amount of evaluations of NHC compensates this, obtaining a far better performance, especially for larger instances.

\begin{figure}
\centering
\includegraphics[width=0.46\textwidth]{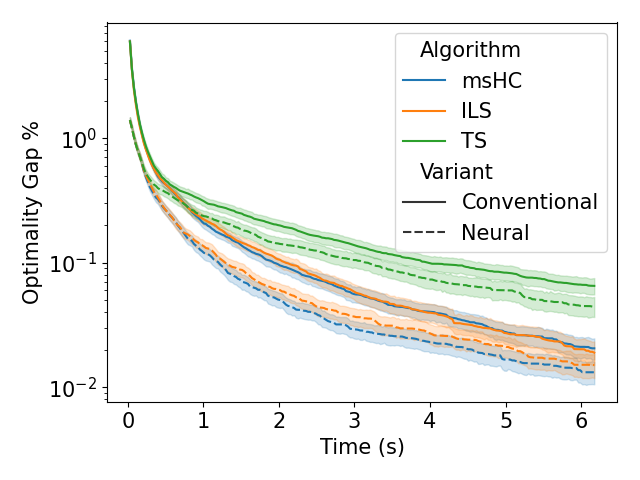}
\caption{Average Optimality gap \% of 512 executions solving instances of $N=20$. The x-axis denotes the computation time.}
\label{fig:performance_time}
\end{figure}

In fact, as can be seen in Fig. \ref{fig:performance_time}, NI based heuristics outperform conventional heuristics under the time-criterion. Neural variants obtain a major advantage in the beginning of the execution, where the decrease in the optimality gap is steeper.
Note that, in order to fairly compare with conventional search strategies, we do not make use of a batch of instances for the model inference, instead, we only feed a unique instance simultaneously. Otherwise, NHC would greatly benefit from using a batch of instances, as done in training, due to parallelization.

\subsubsection{Training time}
When evaluating a learning framework, apart from the inference time, considering training time is also a crucial factor. While computation time for inference is well-researched in the literature, training time is often given less attention. This can pose a challenge when, for example, a model must be trained very frequently, and it is applied to solve a few instances each time. RL frameworks typically require a large amount of computation time for training, which can be expensive in terms of resources and time. If a pre-trained model is not available, conventional methods may be more suitable for solving a single instance since training time is much longer than the execution time required by conventional methods. However, if a pre-trained model is accessible through open-source platforms or from prior practices, and generalizes well to the specific instance being solved, practitioners can still benefit from using the model.

\begin{figure}[h!]
    \centering
    \includegraphics[width=0.46\textwidth]{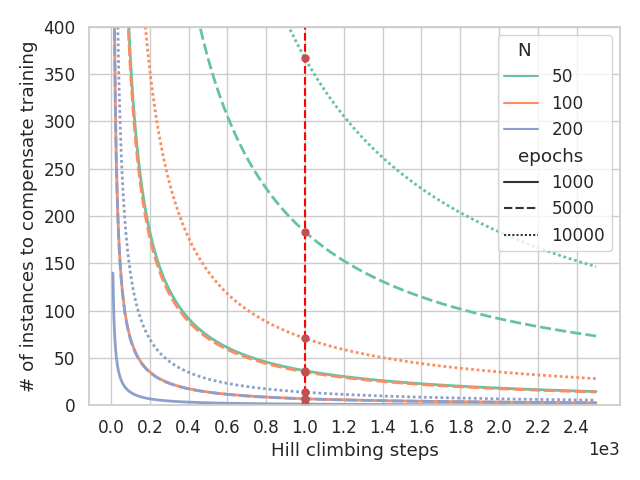}
    \caption{The plot illustrates the minimum number of instances to be solved in order to compensate the training of the model, for each execution time. Trained model sizes are differentiated with different color, while different number of epochs have different line styles (5000 epochs are used in the paper).}
    \label{fig:training_time}
\end{figure}

To analyse the point where it becomes cost effective to train and use a neural method, the following experiment is conducted. We assume that the practitioner does not possess any pre-trained model. It requires the practitioner to weight the cost of training a neural model against the cost of using traditional techniques (inference time) while also considering the expected performance improvement that the neural model can provide. For illustrative purposes, the total time required to complete an execution of the conventional hill climbing and the neural hill climbing are defined in Eqs. \ref{eq:conv} and \ref{eq:ni}, where $y_{conv}$ and $y_{ni}$ are the total time, $t_{train}$, $t_{infer}$ and $t_{neigh}$ denote the training and inference time of the NI model and the time required by the conventional SAHC to evaluate an entire neighborhood, respectively.
\begin{equation}
 \label{eq:conv}
 y_{conv} = t_{neigh} * n_{steps}
\end{equation}
\begin{equation}
 \label{eq:ni}
 y_{ni} = t_{train} + t_{infer} * n_{steps}
\end{equation}
Let $T$ be the number of steps performed in an execution, Eq. \ref{eq:comp} defines the minimum number of instances that need to be solved, to compensate for the training time of the NI model.
\begin{equation}
 \label{eq:comp}
 n_{comp} \geq \left\lceil \frac{t_{train}}{T * (t_{neigh} - t_{infer})} \right\rceil
\end{equation}
Based on our implementation and hardware used (an RTX A5000 GPU and Intel Xeon X5650 CPUs), the conventional SAHC requires an average of 0.40s, 2.06s and 10.41s to explore an entire neighborhood of the PRP of size 50, 100 and 200, respectively. 
The inference time of the NI model is 4.6ms, 13ms and 48ms for PRP instances of size 50, 100 and 200, respectively. 
Considering these values, Fig. \ref{fig:training_time} depicts, for these three problem sizes and three training budgets, the variation in the minimum number of instances that need to be solved to compensate for the training time. The red vertical line represents the conventional stopping criteria (1000$N^2$ evaluations, $T=1000$) used in the literature  \cite{santucci2020using}, and the red dots depict the intersection with different curves. For the training duration of 5000 epochs used in this paper, the minimum number of instances that need to be solved are 183, 35, and 7 for sizes 50, 100, and 200, respectively. 
It is noteworthy that, as opposed to neural methods, the conventional methods require a search in the quadratic neighborhood, and consequently, the difference in execution times between these two approaches becomes more pronounced with larger instance sizes and the training time gets compensated with fewer instances. Therefore, the advantage provided by utilizing the NI model increases dramatically as the instance size increases.

\section{Extension to other problems}
\label{transferability}

In the previous section we have approached the PRP as an illustrative problem where the instance information is exclusively stored in the edges of the graph-representation of the problem. Nevertheless, the proposed model is not exclusive to the PRP, and can be extended to any problem that falls in this category.

In fact, the proposed NI model can be used for other combinatorial problems performing only some minor changes. 
The main change the practitioner needs to perform resides in the  edge-feature selection. 
The NI model uses edges as the main message-passing elements. Extracting instance stationary information and dynamic solution information from edges may not be trivial for every problem. Apart from that, the practitioner needs to select the most efficient operator for the problem at hand.

To illustrate the process, we give examples on how to adapt the model to solve two different problems: the Travelling Salesman Problem (TSP) \cite{applegate2011travelling} and the 
2-partition balanced Graph Partitioning Problem (GPP) \cite{andreev2004balanced}. The reader is headed to the corresponding papers for detailed information about these problems.

\subsection{Traveling Salesman Problem}
Given a set of $n$ cities and their coordinates in a two-dimensional space $s=\{\mathbf{c_i}\}_{i=1}^n$ where each $\mathbf{c_i} \in \mathbb{R}^{2}$, the TSP consists of finding a permutation $\omega$ that orders the cities in a tour that visits each city once and has the minimum length. Formally, the TSP can be defined as a fully connected graph where the nodes represent the cities and weighted edges denote the pairwise distances or costs between cities. 

In the TSP, as opposed to the PRP, certain information is contained in the nodes, namely, the city coordinates. Meanwhile, edge features ($\mathbf{x_{ij}} \in \mathbb{R}^{2}$) are obtained directly by the edge weights in a similar way as done for the PRP: the first dimension in $\mathbf{x_{ij}}$ denotes the distance if the edge is part of the current solution (cities i and j are consecutively visited) and zero otherwise. Similarly, the second dimension is set to non-zero (distance) for the edges that are not part of the solution.

Regarding the operator, as seen in recent works \cite{ma2021learning, wu2021learning}, the 2-opt operator is better suited for routing problems than the insert or the swap operators. However, as done in Fig. \ref{fig:operator}, a set of candidate operators should be considered and evaluated for a few training epochs, in order to select the best one. In fact, in the experiments we opted to use the insert operation since it performs better than 2-opt.

\subsection{2-Partition Balanced Graph Partitioning Problem}
Given a graph $G(N, E, b)$ where $N$ denotes the set of nodes and $E$ the set of edges with weights $b$, the 2-partition balanced GPP consists of finding a 2-partition of $N$, where the number of nodes is balanced among the sets, and which minimizes the sum of the weights of edges going from the nodes of one partition to the nodes in the other.

Regarding the feature extraction, an identity vector is used as node feature, as in the previous cases. Edge features $x_{ij} \in \mathbb{R}^{2}$ are again obtained from the edge weights. $x_{ij} = (b_{ij}, 0)$ when edge $(i,j)$ belongs to the cut, i.e., it is an edge between clusters, and $x_{ij} = (0, b_{ij})$ if it is not.

Regarding the operator, in this case we propose to use swaps between a pair of items in order to guarantee the balance between the two clusters. Even though the model may choose a pair of items in the same cluster, the solution would not change and thus the improvement will be null, forcing the model to pay attention to pairs of items that belong to different clusters.

\subsection{Preliminary results on TSP and GPP}

We evaluated the application of the NI model in the one-step scenario for the described problems using 2000 instances. For the TSP we used instances of 20 cities, placed uniformly at random in the unit square. For the GPP we created random graphs with 20 nodes and 50\% of connectivity with weights sampled from a uniform distribution between $(0, 1)$.

\begin{figure}
\centering
\includegraphics[width=0.45\textwidth]{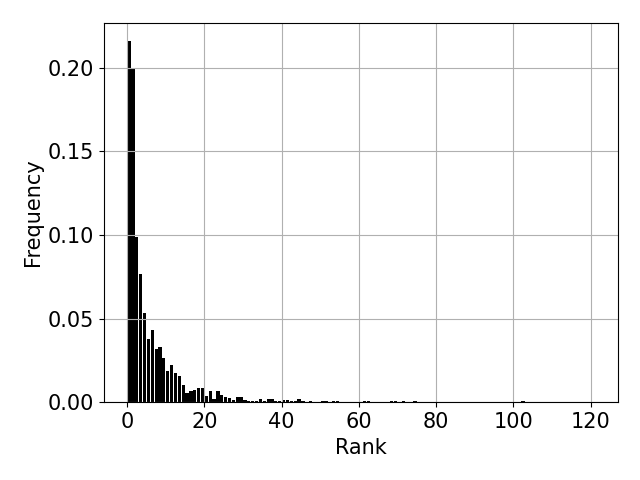}
\caption{TSP. Histogram showing the rankings (x-axis) of the action selected by the model among all the possible actions.}
\label{fig:tsp}
\end{figure}

\begin{figure}
\centering
\includegraphics[width=0.45\textwidth]{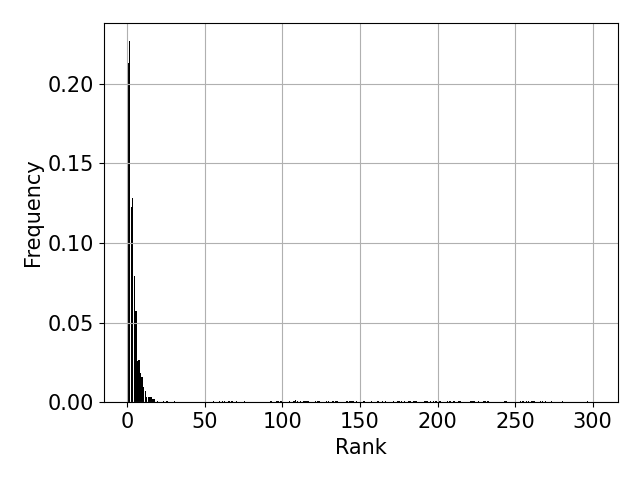}
\caption{GPP. Histogram showing the rankings (x-axis) of the action selected by the model among all the possible actions.}
\label{fig:gpp}
\end{figure}

We trained a NI model for each problem following the setup described in Section \ref{setup}. In the TSP, the NI model selects, on average, the action ranked $5^th$ out of 361 (see Fig. \ref{fig:tsp} for further details on the ranks of the selected actions). In the GPP, the NI model selects, on average, the action ranked $10^th$, however, among the 2000 instances, the model selected an invalid swap (swap between two items in the same cluster) in 89 cases (4\%) and thus, some actions seem to be ranked worse than usual (see Fig. \ref{fig:gpp}). If we masked the invalid moves, the NI selects, on average, the action ranked 3rd.

\section{Conclusion and Future Work}
\label{conclusion}
This paper presents a Neural Improvement model for graph-based problems which, given a candidate solution, is able to propose a pairwise modification that (almost) always generates a new better solution. 
We have experimentally demonstrated that the NI module could replace traditional local-search strategies, since it requires less computational effort to obtain similar results, it is more flexible, and can efficiently guide a variety of hill-climbing algorithms.
This work has major implications for most of the state-of-the-art metaheuristics used to solve combinatorial optimization problems, as they commonly include conventional local search procedures. 


The research presented in this paper represents a promising avenue for future investigation. In particular, we emphasize the importance of investigating different training strategies, given that the performance of the model is heavily influenced by the instances used in training. To this end, we suggest exploring advanced curriculum learning strategies that intelligently select training instances, as well as real-world instance generators that would allow to generate a broader range of instance closer to the target distribution.

Although the model has shown good performance, local search can be prone to becoming trapped in local optima, as is the case with conventional approaches. We think that a more advanced model that can overcome this limitation should be developed. In particular, we suggest incorporating strategies such as curiosity-driven learning \cite{pathak2017curiosity} and memory-based learning \cite{badia2020agent57} to improve the exploration of the NI model and prevent the model from revisiting previously explored states. By taking these steps, we believe that the potential of the NI model can be fully realized, opening up exciting opportunities for future research in this area.

{\appendices
\section{Hyperparameter and architecture selection}
\label{app1}
In order to select the best model, we have experimented with several options. As an encoder, we have evaluated two of the most popular options, Graph Attention Networks (GAT) \cite{velickovic2017graph} and Anisotropic Graph Neural Networks (GNN) \cite{dwivedi2020benchmarking}. For the GAT encoder to be able to consider edges, we have used the edge-featured GAT \cite{chen2021edge}. As the decoder, we have tested a Multi-Head Attention mechanism (MHA) \cite{vaswani2017attention}, which performs a self-attention pass over the edge embeddings before generating the output probabilities, and a simple Multi-Layer Perceptron (MLP), which directly produces the output probabilities without any attention.

For evaluating different encoder and decoder architectures, we have trained three different models for 100 epochs and evaluated on a set of test instances every epoch. Fig. \ref{fig:architecture_search} shows the reward curves for three different architectures: (1) GAT encoder and MHA decoder, (2) GNN encoder and MHA decoder, and (3) GNN encoder and MLP decoder. As shown in the figure, GNN encoder obtains better results compared to GAT, while there is not such difference between decoders. Moreover, GAT encoder is time consuming (4 times slower than GNN). As a result, we have chosen GNN as the encoder. Among decoders, we have selected the MLP, since it is slightly faster and more stable than MHA.

\begin{figure}[h!]
    \centering
    \includegraphics[width=0.49\textwidth]{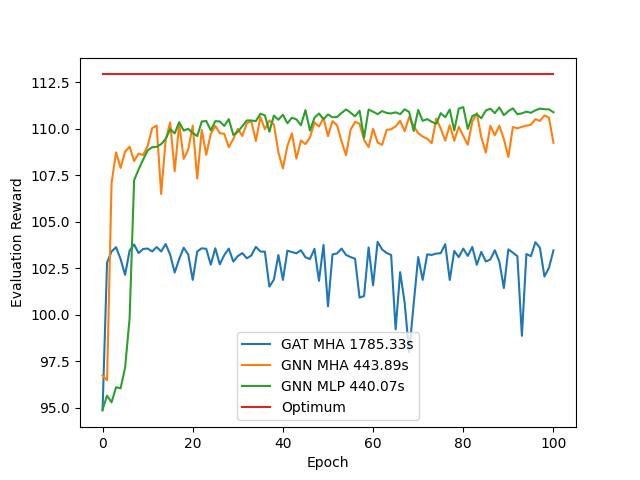}
    \caption{Evaluation reward during training for different model architectures. Three different architectures are tested: (1) GAT encoder and MHA decoder, (2) GNN encoder and MHA decoder, and (3) GNN encoder and MLP decoder.}
    \label{fig:architecture_search}
\end{figure}

Table \ref{hyperparams} provides a description of the hyperparameters used in the proposed model. We have set the embedding dimension to 128, which is a commonly used power-of-two value. Through experimentation, we determined that this value provides the best trade-off between performance and computational efficiency. Additionally, we have set the number of layers to $L=3$. Fewer layers are insufficient for correctly encoding the graph structure, while larger GNNs may encounter the issue of oversmoothing.
Parameters (3-7) have been set with values reported with similar architectures \cite{bello2016neural, joshi2020learning, ma2021learning}.
The discount factor is also a key aspect in the reward engineering process. We found out that using a low discount factor ($\gamma = 0.1$) yields better results.
The episode length, denoted as $T$, specifies how frequently the reward is updated. A model trained using shorter episodes, such as $T=1$, tends to learn policies similar to Steepest-Ascent Hill Climbing. In contrast, longer episodes do not necessarily promote greedy behavior; instead, they allow the model to optimize for long-term rewards. The parameter $K_{max}$ denotes the maximum number of steps the model can take without improving the best discovered reward. Increasing $K_{max}$ can provide the model with more opportunities to explore high-quality solutions. However, excessively large values are computationally expensive and unnecessary.

\begin{table}
 \caption{Training and model hyperparameters}
\label{hyperparams}
  \centering
\begin{tabular}{llrr}
\toprule

&Hyperparameter & Symbol  & Value \\ \midrule
1 & Embedding dimension & $d$ & 128 \\
2 & Number of encoding layers & $L$ & 3 \\
3 & Tanh clipping & C & 10 \\
4 & Gradient norm clipping & - & 1 \\
5 & Learning rate & $\alpha$ & 1e-4 \\
6 & Batch size & $B$ & 64 \\
7 & Number of epochs & $n_{epochs}$ & 5000 \\ 

8 & Discount factor & $\gamma$ & 0.1 \\
9 & Episode length & $T$ & 20 \\
10 & Max num. of consecutive & & \\ 
& non-improving moves & $K_{max}$ & 5 \\
\end{tabular}
\end{table}%

\section {Generalization to Other Instance Types}
\label{app2}
\begin{table}
  \caption{Comparison of the performance of NHC and the best conventional HC (BFHC) in different instances of the XLOLIB benchmark. The shown values are average gaps to the optimal value taken from 100 executions.}
  \label{tab:lolib}%
  \centering
\begin{tabular}{lrrlrr}
\toprule
Instance & NHC & BFHC & Instance & NHC & BFHC \\ \midrule
N-t65b11xx\_250  & \textbf{3.14}\% & 3.66\%  & N-tiw56n62\_250 & \textbf{3.57}\% & 4.42\% \\
N-stabu3\_250    & \textbf{3.34}\% & 3.97\%  & N-be75eec\_250  & \textbf{3.85}\% & 3.91\% \\
N-t74d11xx\_250  & \textbf{3.08}\% & 3.61\%  & N-tiw56n66\_250 & \textbf{2.93}\% & 3.53\% \\
N-be75oi\_250    & \textbf{3.12}\% & 3.32\%  & N-t70d11xn\_250 & \textbf{2.75}\% & 3.68\% \\
N-t59n11xx\_250  & \textbf{2.89}\% & 4.89\%  & N-t75k11xx\_250 & \textbf{3.37}\% & 3.79\% \\
N-tiw56r58\_250  & \textbf{3.34}\% & 3.78\%  & N-tiw56r54\_250 & \textbf{3.16}\% & 3.40\% \\
N-t59f11xx\_250  & \textbf{3.36}\% & 3.77\%  & N-t70n11xx\_250 & \textbf{4.08}\% & 4.42\% \\
N-tiw56n54\_250  & \textbf{2.78}\% & 3.65\%  & N-tiw56r72\_250 & \textbf{3.17}\% & 3.59\% \\
N-t69r11xx\_250  & \textbf{3.70}\% & 3.86\%  & N-t65f11xx\_250 & \textbf{3.82}\% & 3.97\% \\
N-t75d11xx\_250  & \textbf{3.06}\% & 3.68\%  & N-t70f11xx\_250 & \textbf{4.11}\% & 4.27\% \\
N-be75np\_250    & \textbf{3.39}\% & 3.62\%  & N-tiw56n67\_250 & \textbf{3.45}\% & 3.79\% \\
N-t75n11xx\_250  & \textbf{3.73}\% & 3.99\%  & N-tiw56r66\_250 & \textbf{3.97}\% & 4.20\% \\
N-t75e11xx\_250  & \textbf{3.64}\% & 4.07\%  & N-t65d11xx\_250 & \textbf{3.43}\% & 4.12\% \\
N-tiw56n58\_250  & \textbf{3.57}\% & 3.68\%  & N-stabu1\_250   & \textbf{3.00}\% & 3.72\% \\
N-stabu2\_250    & \textbf{2.95}\% & 3.46\%  & N-tiw56n72\_250 & \textbf{2.89}\% & 3.79\% \\
N-tiw56r67\_250  & \textbf{2.65}\% & 3.70\%  & N-t70b11xx\_250 & 3.52\% & \textbf{3.50}\% \\
N-t70l11xx\_250  & \textbf{4.12}\% & 4.71\%  & N-t59d11xx\_250 & \textbf{3.37}\% & 3.80\% \\
N-t65n11xx\_250  & \textbf{3.65}\% & 4.68\%  & N-t70d11xx\_250 & \textbf{3.51}\% & 4.38\% \\
N-t65l11xx\_250  & \textbf{2.93}\% & 4.01\%  & N-be75tot\_250  & \textbf{3.53}\% & 3.96\% \\
N-t59b11xx\_250  & \textbf{3.75}\% & 4.37\%  &                 &        &        \\
\end{tabular}
\end{table}%

The proposed model has been designed for the Preference Ranking Problem (PRP); however, since there is no any widely used benchmark in the PRP, we have opted for the Linear Ordering Problem (LOP)\cite{ceberio2015linear}, a classical combinatorial optimization problem that can be seen as a mathematical formulation of the PRP. To further validate the proposal in real-world data, and demonstrate its ability to generalize to different instance distributions, the NHC method has been evaluated in the most popular library for the LOP, the LOLIB\cite{reinelt2002linear}. Particularly, we selected the most challenging instance type, the XLOLIB of size 250. Table \ref{tab:lolib} summarizes the average performance gap to the best known value obtained by \cite{lugo2021diversity} in 100 executions limited by 1 minute. NHC obtains better average gap for 38 out of 39 instances compared to the best performing conventional hill climbing method (Best First Hill Climbing).
}

\section*{Acknowledgments}
Andoni Irazusta Garmendia acknowledges a predoctoral grant from the Basque Government (ref. PRE\_2020\_1\_0023). This work has been partially supported by the Research Groups 2022-2024 (IT1504-22) and the Elkartek Program (KK-2022/00106, SIGZE, KK- 2021/00065) from the Basque Government, the PID2019-104933GB-10 and PID2019-106453GA-I00/AEI/10.13039/501100011033 research projects from the Spanish Ministry of Science. Finally, we acknowledge the support of NVIDIA Corporation with the donation of a RTX A5000 GPU used for this work.

\vfill


\begin{thebibliography}{1}
\bibliographystyle{IEEEtran}

\bibitem{paschos2014applications}
V. T. Paschos. "Applications of combinatorial optimization," in \textit{ John Wiley \& Sons} vol. 3, 2014.

\bibitem{naseri2020application}
G. Naseri, M. A. Koffas. "Application of combinatorial optimization strategies in synthetic biology" in \textit{Nature communications}, vol. 11, no. 1, 2020.

\bibitem{garey1979computers}
M. R. Garey, D. S. Johnson. "Computers and intractability," \textit{A Guide to the Theory of NP-Completeness.} 1979.

\bibitem{chen2018foad}
J. Chen, C. Liu, M. Tomizuka. "Foad: Fast optimization-based autonomous driving motion planner," in \textit{2018 Annual American Control Conference}, pp. 4725-4732, 2018.

\bibitem{talbi2021machine}
E. G. Talbi. "Machine learning into metaheuristics: A survey and taxonomy," in \textit{ACM Computing Surveys}, vol. 54, no. 6, pp. 1-32, 2021.

\bibitem{mazyavkina2021reinforcement}
N. Mazyavkina, S. Sviridov, S. Ivanov, E. Burnaev. "Reinforcement learning for combinatorial optimization: A survey," in \textit{Computers \& Operations Research}, p. 134, 2021.

\bibitem{bengio2021machine}
Y. Bengio, A. Lodi, A. Prouvost. "Machine learning for combinatorial optimization: a methodological tour d’horizon," in \textit{European Journal of Operational Research}, vol. 290, no. 2, pp. 405-421, 2021.

\bibitem{bello2016neural}
I. Bello, H. Pham, Q. V. Le, M. Norouzi, S. Bengio. "Neural combinatorial optimization with reinforcement learning," in \textit{arXiv preprint arXiv:1611.09940}, 2016.

\bibitem{kool2018attention}
W. Kool, H. Van Hoof, M. Welling. "Attention, learn to solve routing problems!," in \textit{arXiv preprint arXiv:1803.08475}, 2018.

\bibitem{kwon2020pomo}
Y. D. Kwon, J. Choo, B. Kim, I. Yoon, Y. Gwon, S. Min. "Pomo: Policy optimization with multiple optima for reinforcement learning," in \textit{Advances in Neural Information Processing Systems}, vol. 33, pp. 21188-21198, 2020.

\bibitem{chen2019learning}
X. Chen, Y. Tian. "Learning to perform local rewriting for combinatorial optimization," in \textit{Advances in Neural Information Processing Systems}, p. 32, 2019.

\bibitem{lu2019learning}
H. Lu, X. Zhang, S. Yang. "A learning-based iterative method for solving vehicle routing problems," in \textit{International conference on learning representations}, 2019.

\bibitem{wu2021learning}
Y. Wu, W. Song, Z. Cao, J. Zhang, A. Lim. "Learning Improvement Heuristics for Solving Routing Problems," in \textit{IEEE transactions on neural networks and learning systems}, 2021.

\bibitem{accorsi2022guidelines}
L. Accorsi, A. Lodi, D. Vigo. "Guidelines for the computational testing of machine learning approaches to vehicle routing problems," in \textit{Operations Research Letters}, vol. 50, no. 2, pp. 229-234, 2022.

\bibitem{garmendia2022neural}
A. I. Garmendia, J. Ceberio, A. Mendiburu. "Neural Combinatorial Optimization: a New Player in the Field," in \textit{arXiv preprint arXiv:2205.01356}, 2022.

\bibitem{heckel2018approximate}
R. Heckel, M. Simchowitz, K. Ramchandran, M. Wainwright. "Approximate ranking from pairwise comparisons," in \textit{International Conference on Artificial Intelligence and Statistics}, pp. 1057-1066, 2018.

\bibitem{andreev2004balanced}
K. Andreev, H. Räcke. "Balanced graph partitioning," in \textit{Proceedings of the sixteenth annual ACM symposium on Parallelism in algorithms and architectures}, pp. 120-124, 2004.

\bibitem{applegate2011travelling}
D. L. Applegate, R. E. Bixby, V. Chvatal, W. J. Cook. "The traveling salesman problem: a computational study," in \textit{Princeton university press}, 2011.

\bibitem{hopfield82}
J. J. Hopfield, D. W. Tank. “Neural computation of decisions in optimization problems," in \textit{Biological cybernetics}, vol. 52, no. 3, pp. 141-152, 1985.

\bibitem{vinyals2015pointer}
O. Vinyals, M. Fortunato, N. Jaitly. "Pointer networks," in \textit{Advances in Neural Information Processing Systems}, vol. 28, pp. 2692-2700, 2015.

\bibitem{cappart2021combinatorial}
Q. Cappart, D. Chételat, E. Khalil, A. Lodi, C. Morris, P. Veličković. "Combinatorial optimization and reasoning with graph neural networks," in \textit{arXiv preprint arXiv:2102.09544}, 2021.

\bibitem{joshi2020learning}
C. K. Joshi, Q. Cappart, L. M. Rousseau,T. Laurent, X. Bresson. "Learning TSP requires rethinking generalization," in \textit{arXiv preprint arXiv:2006.07054}, 2020.

\bibitem{hottung2019neural}
A. Hottung, K. Tierney. "Neural large neighborhood search for the capacitated vehicle routing problem" in \textit{arXiv preprint arXiv:1911.09539}, 2019.

\bibitem{da2021learning}
P. da Costa, J. Rhuggenaath, Y. Zhang, A. Akcay, U. Kaymak. "Learning 2-opt heuristics for routing problems via deep reinforcement learning," in \textit{SN Computer Science}, vol. 2, pp. 1-16, 2021.

\bibitem{falkner2023learning}
J. K. Falkner, D. Thyssens, A. Bdeir, L. Schmidt-Thieme. "Learning to Control Local Search for Combinatorial Optimization," in \textit{Machine Learning and Knowledge Discovery in Databases: European Conference}, pp. 361-376, 2023.

\bibitem{ma2021learning}
Y. Ma, J. Li, Z. Cao, W. Song, L. Zhang, Z. Chen, J. Tang. "Learning to iteratively solve routing problems with dual-aspect collaborative transformer," in \textit{Advances in Neural Information Processing Systems}, vol. 34, pp. 11096-11107, 2021.

\bibitem{joshi2019efficient}
C. K. Joshi, T. Laurent and X. Bresson. "An efficient graph convolutional network technique for the travelling salesman problem," in \textit{arXiv preprint arXiv:1906.01227}, 2019.

\bibitem{fu2021generalize}
Z. H. Fu, K. B. Qiu, H. Zha, H. "Generalize a small pre-trained model to arbitrarily large tsp instances," in \textit{Proceedings of the AAAI Conference on Artificial Intelligence}, vol. 35, no. 8, pp. 7474-7482, 2021.

\bibitem{tromble2009learning}
R. Tromble, J. Eisner. "Learning linear ordering problems for better translation," in \textit{Proceedings of the 2009 conference on empirical methods in natural language processing}, pp. 1007-1016, 2009.

\bibitem{leontief1986input}
W. Leontief. "Input-output economics," in \textit{Oxford University Press}, 1986.

\bibitem{achatz2006corruption}
H. Achatz, P. Kleinschmidt, J. Lambsdorff. "Der corruption perceptions index und das linear ordering problem," in \textit{ORNews}, vol. 26, pp. 10-12, 2006.

\bibitem{anderson2019rankability}
P. Anderson, T. Chartier, A. Langville. "The rankability of data," in \textit{SIAM Journal on Mathematics of Data Science}, vol. 1 no. 1, pp. 121-143, 2019.

\bibitem{cameron2021linear}
T. Cameron, S. Charmot, J. Pulaj. "On the linear ordering problem and the rankability of data," in \textit{arXiv preprint arXiv:2104.05816}, 2021.

\bibitem{shah2013case}
N. B. Shah, J. K. Bradley, A. Parekh, N. Wainwright, K. Ramchandran. "A case for ordinal peer-evaluation in MOOCs," in \textit{NIPS workshop on data driven education}, vol. 15, p. 67, 2013.

\bibitem{kwon2021matrix}
Y. D. Kwon, J. Choo, I. Yoon, M. Park, D. Park, Y. Gwon. "Matrix encoding networks for neural combinatorial optimization," in \textit{Advances in Neural Information Processing Systems}, vol. 34, pp. 5138-5149, 2021.


\bibitem{williams1992simple}
R. J. Williams. "Simple statistical gradient-following algorithms for connectionist reinforcement learning," in \textit{Machine learning}, vol. 8, no. 3, pp. 229-256, 1992.

\bibitem{bengio2009curriculum}
Y. Bengio, J. Louradour, R. Collobert, J. Weston. "Curriculum learning," in \textit{Proceedings of the 26th annual international conference on machine learning}, pp. 41-48, 2009.

\bibitem{laguna1999intensification}
M. Laguna, R. Marti, V. Campos. "Intensification and diversification with elite tabu search solutions for the linear ordering problem," in \textit{Computers \& Operations Research}, vol. 26, no. 12, pp. 1217-1230, 1999.

\bibitem{blum2003metaheuristics}
C. Blum, A. Roli. "Metaheuristics in combinatorial optimization: Overview and conceptual comparison," in \textit{ACM computing surveys}, vol. 35, no. 3, pp. 268-308, 2003.

\bibitem{glover1993user}
F. Glover, E. Taillard. "A user's guide to tabu search," in \textit{Annals of operations research}, vol. 41, no. 1, pp. 1-28, 1993.

\bibitem{lourenco2019iterated}
H. R. Lourenço, O. C. Martin, T. Stützle. "Iterated local search: Framework and applications," in \textit{Handbook of metaheuristics}, pp. 129-168, 2019.

\bibitem{Becker1967}
O. Becker. "Das Helmstädtersche Reihenfolgeproblem—die Effizienz verschiedener Näherungsverfahren," in \textit{Computer uses in the Social Sciences}, 1967.

\bibitem{lugo2021diversity}
L. Lugo, C. Segura, G. Miranda. "A diversity-aware memetic algorithm for the linear ordering Problem," in \textit{Memetic Computing}, pp. 1-15, 2021.

\bibitem{santucci2020using}
V. Santucci, J. Ceberio. "Using pairwise precedences for solving the linear ordering problem," in\textit{Applied Soft Computing}, vol. 87, 2020.

\bibitem{pathak2017curiosity}
D. Pathak, P. Agrawal, A. A. Efros, T. Darrell. "Curiosity-driven exploration by self-supervised prediction," in \textit{International conference on machine learning}, pp. 2778-2787, 2017.

\bibitem{badia2020agent57}
A. P. Badia, B. Piot, S. Kapturowski, P. Sprechmann, A. Vitvitskyi, Z. D. Guo, C. Blundell. "Agent57: Outperforming the atari human benchmark" in \textit{International Conference on Machine Learning}, pp. 507-517, 2020.

\bibitem{velickovic2017graph}
P. Veličković, G. Cucurull, A. Casanova, A. Romero, P. Lio, Y. Bengio, Y. "Graph attention networks". arXiv preprint arXiv:1710.10903. 2017.

\bibitem{dwivedi2020benchmarking}
V. P. Dwivedi, C. K. Joshi, T. Laurent, Y. Bengio, X. Bresson, X. "Benchmarking graph neural networks". 2020.

\bibitem{chen2021edge}
J. Chen, H. Chen. "Edge-featured graph attention network". arXiv preprint arXiv:2101.07671. 2021.

\bibitem{vaswani2017attention}
A. Vaswani, N. Shazeer, N. Parmar, J. Uszkoreit, L. Jones, A. N. Gomez, ... , I. Polosukhin. "Attention is all you need". in \textit{Advances in neural information processing systems, 30}. 2017.

\bibitem{ceberio2015linear}
J. Ceberio, A. Mendiburu, J. A. Lozano. "The linear ordering problem revisited" in \textit{European Journal of Operational Research}, vol. 241, no. 3, pp. 686-696, 2015.

\bibitem{reinelt2002linear}
G. Reinelt. "Linear ordering library (LOLIB)". University of Heidelberg, 2002.

\end{thebibliography}
\end{document}